\pdfoutput=1

\documentclass[11pt]{article}

\usepackage{acl}

\usepackage{times}
\usepackage{latexsym}

\usepackage[T1]{fontenc}

\usepackage[utf8]{inputenc}

\usepackage{microtype}

\usepackage{tabularx}
\usepackage{tablefootnote}
\usepackage{amsmath}
\usepackage{soul}

\newcolumntype{L}[1]{>{\raggedright\let\newline\\\arraybackslash\hspace{0pt}}m{#1}}
\newcolumntype{C}[1]{>{\centering\let\newline\\\arraybackslash\hspace{0pt}}m{#1}}
\newcolumntype{R}[1]{>{\raggedleft\let\newline\\\arraybackslash\hspace{0pt}}m{#1}}

\usepackage{tikz}
\usetikzlibrary{calc,spy,shapes}
\usetikzlibrary{arrows,decorations.pathmorphing,fit,positioning,patterns}
\usepackage{pgfplots}
\pgfplotsset{compat=1.16}

\usepackage{pdfpages}

\usepackage{helvet}
\usepackage[eulergreek]{sansmath}

\definecolor{lancsred}{RGB}{178,14,16}

\title{Understanding who uses Reddit: Profiling individuals with a self-reported bipolar disorder diagnosis}

\author{Glorianna Jagfeld\textsuperscript{$\star$}, Fiona Lobban\textsuperscript{$\star$}, Paul Rayson\textsuperscript{$\bigtriangledown$}, Steven H. Jones\textsuperscript{$\star$} \\ 
  \textsuperscript{$\star$}Spectrum Centre for Mental Health Research\\
    \textsuperscript{$\bigtriangledown$}School of Computing and Communications \\
  Lancaster University, United Kingdom \\
  \texttt{\{g.jagfeld, f.lobban, p.rayson, s.jones7\}@lancaster.ac.uk}
}

\begin{document}
\maketitle
\begin{abstract}
Recently, research on mental health conditions using public online data, including Reddit, has surged in NLP and health research but has not reported user characteristics, which are important to judge generalisability of findings.
This paper shows how existing NLP methods can yield information on clinical, demographic, and identity characteristics of almost 20K Reddit users who self-report a bipolar disorder diagnosis.
This population consists of slightly more feminine- than masculine-gendered mainly young or middle-aged US-based adults who often report additional mental health diagnoses, which is compared with general Reddit statistics and epidemiological studies. 
Additionally, this paper carefully evaluates all methods and discusses ethical issues.
\end{abstract}

\section{Introduction and related work}
People who experience extreme 
mood states that interfere with their functioning, meet the criteria for bipolar disorder~(BD) according to the diagnostic manuals Diagnostic and Statistical Manual of Mental Disorders~(DSM)~\citep{DSM-5_APA_13} and International Classification of Diseases~(ICD)~\citep{ICD-11_WHO_18}.
DSM and ICD operationalise extreme mood states in terms of major depressive episodes, \lq almost daily depressed mood or diminished interest in activities with additional symptoms for at least 14 days\rq~\citep{ICD-11_WHO_18} and (hypo-)manic episodes, \lq a distinct period of abnormally and persistently elevated, expansive, or irritable mood and abnormally and persistently increased goal-directed activity or energy\rq~ that lasts at least seven (four) days~\citep[p.~124]{DSM-5_APA_13}.

DSM and ICD distinguish several BD subtypes based on the lifetime frequency and intensity of (hypo-)manic and depressed episodes. 
The only requirement for a diagnosis of bipolar I disorder (BD-I) is at least one lifetime manic episode, whereas bipolar II disorder (BD-II) requires at least one hypomanic and one major depressive episode~\citep[pp.~126,~132]{DSM-5_APA_13}.
Cyclothymic disorder applies to numerous periods of hypomanic and depressive symptoms during at least two years that do not meet criteria for hypomanic or major depressive episodes~\citep[p.~139]{DSM-5_APA_13}.

Bipolar mood episodes are often recurring~\citep{Treuer2010,Gignac2015}, so many individuals living with BD require life-long treatment~\citep{Goodwin2016b} and have a heightened suicide risk~\citep{Novick2010}.
However, characteristics and outcomes of people meeting BD criteria are diverse, with some living well, 
~\citep[e.g.,][]{LivingLIifeYouWant_Warwick_19}
and even functioning on a high level~\citep{SocialOccupationalFunctioningBD_Akers_19}.

\subsection{Online forums as research data source}
Online forums have become an increasingly attractive source for research data, enabling non-reactive data collection, where researchers do not influence data creation, at large scale~\citep{Fielding2016}.
Natural language processing~(NLP) research in this area has focused on predicting people at risk of BD~\citep{QuantifyingMentalHealthTwitter_Coppersmith_14,SMHD_Cohan_18,BDPrediction_Sekulic_18}.
Health researchers have explored the lived experience of BD with qualitative analyses of online posts~\citep{Mandla2017,BDReddit_Sahota_19}.
Unlike in clinical studies, usually little or no demographic information is available for online forum users, so it is unclear to what populations these results generalise~\citep{Ruths2014}.
For example, language differences between Twitter users with self-reported Major depressive disorder (MDD) or Post-traumatic stress disorder (PTSD) correlated highly with their personality and demographic characteristics~\citep{PersonalityAgeGenderTweetingMentaIllness_Preotiuc-Pietro_15}.
So it is unclear whether these findings really indicate mental health~(MH) diagnoses or other user characteristics.

\subsection{The online discussion forum Reddit}
Besides MH-specific platforms~\citep{Kramer2004,BipolarDisorderOnlineForum_Vayreda_09,OnlineForumsBipolarDisorder_Bauer_13,DiscussionPagesBD_Latalova_14,OnlineForumBipolarDisorder_Poole_15,OnlineSupportSystemicFuntionalLinguistics_McDonald_16,Campbell2019}, blogs~\citep{Mandla2017}, and Twitter~\citep{QuantifyingMentalHealthTwitter_Coppersmith_14,Ji2015,Saravia2016,BDTwitter_Budenz_19,BDRecognition_Huang_19}, much recent research of user-generated online content in BD has focused on the international online discussion forum Reddit\footnote{\url{https://www.reddit.com/}}~\citep{LanguageMentalHealthSocialMedia_Gkotsis_16,Gkotsis2017d,SMHD_Cohan_18,BDPrediction_Sekulic_18,BDReddit_Sahota_19,Yoo2019}.

The platform Reddit is among the most visited internet sites worldwide~\citep{AlexaReddit_20}, 
hosting a number of subforums (\lq subreddits\rq ) for general topics as well as interest groups. 
There is a vast and growing amount of BD-related content on Reddit, with more than 50K new posts per month in the four largest BD-related subreddits\footnote{r/bipolar, r/BipolarReddit, r/bipolar2, r/bipolarSOs}.
Anyone can view posts without registration 
and the Reddit API offers free access to all historic posts.
%
Reddit profiles do not provide any user characteristics besides the username and sign-up date in a structured format or comparable to a Twitter bio.
While some surveys provide general information on Reddit users, none of the BD-specific studies looked at particular user characteristics of their sample, which is important~\citep{RedditData_Amaya_19}.

\subsection{Research questions and contributions}
The above considerations motivate our research questions:
What characteristics of Reddit users who disclose a BD diagnosis can be automatically inferred from their public Reddit information 
and how do they compare to general Reddit users and clinical populations?
What are the ethical considerations around determining users' characteristics and ways to minimise potential negative impacts?

This work has two main contributions, both of which may be relevant to different parts of the CLPsych community.
Crucially, the authors are an interdisciplinary team of NLP and clinical psychology researchers, as well as practising clinical psychologists, who regularly consult with people with lived experience of BD in an advisory panel.

First, this paper estimates and discusses clinical, demographic and identity characteristics of Reddit users who self-report a BD diagnosis~(see Figure~\ref{sec:appendix-visual-abstract} for a visual results summary).
This has implications for future BD-focused research on Reddit and helps to contextualise previous work.
Moreover, this information is relevant for clinicians who may want to recommend certain online forums to clients and to clinical researchers interested in recruiting via Reddit.
Second, this work shows how simple rule-based and off-the-shelf state-of-the-art NLP methods can estimate Reddit user characteristics, and carefully discusses ethical considerations and harm-mitigating
ways of doing so.
These findings and discussions apply to other, also non-clinical, subgroups of Reddit users.
The evaluation with manual annotations evaluates published NLP methods in an applied setting.

\begin{table*}[tb]
\centering
\begin{tabular}{lrp{10cm}}
\textbf{Component} & \textbf{Number} & \textbf{Examples}\\
\hline
Inclusion patterns	&145&	As someone with a diagnos*, my recent CONDITION diagnos*, ~~~~~I went to a DOCTOR and got diagnos* \\
CONDITION terms	&92&	Bipolar, manic depression, BD-I, BD-II, cyclothymia\\
DOCTOR terms	&18	&Doctor, pdoc, shrink\\
Exclusion patterns	&74	&Not formally diagnos*, self diagnos*, she’s diagnos*
\end{tabular}
\caption{\label{tab:patterns} Components of patterns to identify English self-reported diagnosis statements; *: wildcard
}
\end{table*}

\section{Methods}

\subsection{User identification}
In this work, the identification of Reddit users with lived experience of BD adapts previous approaches based on self-reported diagnosis statements, e.g., \lq I was diagnosed with BD today\rq~\citep{ADHDToSAD_Coppersmith_15,SMHD_Cohan_18,BDPrediction_Sekulic_18}.
Importantly, this captures self-\textit{reported} diagnoses by a professional and not \textit{self-diagnoses}, which were excluded.
Contrary to existing datasets of Reddit posts by people with a self-reported BD diagnosis, all posts of identified people were retained and not only those unrelated to MH concerns.
This enables subsequent research on the lived experience of people with BD. 
All available Reddit posts (January 05 - March 19) that mentioned \lq diagnosis\rq~  and a BD term (see below) were downloaded from Google BigQuery. 
User account meta-data (id, username, UTC timestamp of sign-up) for all matching posts was retrieved via the Reddit python API praw\footnote{\url{https://github.com/praw-dev/praw}} to remove posts by users who had deleted their profile after creation of the BigQuery tables.
Each of the 170K posts was classified as self-reported diagnosis post after automatically removing quoted content if it met the following criteria adapted from~\citet{SMHD_Cohan_18} (see Table~\ref{tab:patterns} for examples):
\begin{itemize}
\item Contains at least one condition term for BD.
\item Matches at least one inclusion pattern, i.e., BD diagnosis of any type by a professional. 
\item Does not match any exclusion pattern, e.g., self-diagnosis.
\item The distance between at least one condition term and the beginning or end of an inclusion phrase is less than the experimentally determined threshold of 55~characters.
\end{itemize}

Subsequently, all posts  (id, submissions title, text, subreddit, user id, UTC timestamp of time posted) of the 21K~user accounts with at least one self-reported diagnosis post were downloaded via praw.
The first author checked the self-reported diagnosis statements of all accounts with more than 1.5K~submissions or 200K~comments or whose name included \lq bot\rq~ or \lq auto\rq , removing 30 automated user accounts~(bots).
Finally, 960~user accounts with a self-reported psychotic disorder diagnosis were removed because this constitutes an exclusion criterion for BD
~\citep[pp.~126,~134]{DSM-5_APA_13}.

\subsection{User characteristics extraction/inference}
\label{sec:methods-demographics}
Several NLP methods were applied and compared to extract or infer clinical~(MH comorbidities = diagnoses additional to BD), demographic~(age, country of residence), and identity~(gender) characteristics of Reddit users with a self-reported BD diagnosis.
See Appendix~\ref{sec:appendix-method} for more details on the age, country, and gender methods and their previously published performance.
The first and third author manually annotated self-reported BD diagnoses, age, country, and gender for random included users for evaluation.

\subsubsection{Mental health comorbidities}
Frequencies for other self-reported MH diagnoses were obtained by matching all dataset posts against inclusion patterns for other diagnoses, in the same way as for identifying self-reported BD diagnoses.
Condition terms for nine major DSM-5 and ICD-11 diagnoses were extended from~\citet{SMHD_Cohan_18}: Anxiety disorder (Generalised/Social anxiety disorder, Panic disorder), Attention deficit hyperactivity disorder (ADHD), Borderline personality disorder (BPD), MDD, PTSD, Psychotic disorder (Schizophrenia/Schizoaffective disorder), Obsessive compulsive disorder (OCD), Autism spectrum disorder (ASD), and Eating disorder (ED).

\subsubsection{Age}
\label{sec:methods-age}
Two methods to recognise a user's age relative to one of their posts were compared. 
An approximate date of birth was calculated from the post timestamp to then calculate the user's age when posting for the first time and their mean age over all posts. 

\begin{itemize}
\item \textbf{Self-reported}: Reddit users sometimes self-report their age and gender in a bracketed format, e.g. \lq I [17f] just broke up with bf [18m]\lq.
Regular expressions extracted age and gender from such self-reports in submission titles. 
\item \textbf{Language use}: Tigunova's~(\citeyear{HAM_Tigunova_19}) neural network model predicts the age group of users with at least ten posts from their contents and language style.
Training data for this model came from \citet{RedditUserTraits_Tigunova_20} who automatically labelled Reddit users with their self-reported age~(see Appendix~\ref{sec:appendix-method-age+gender}).

\item \textbf{Hybrid}:
The Hybrid method assigns the extracted age from the Self-reported method if available, and otherwise the predicted age from the Language use method because evaluation revealed that the Self-reported method had higher accuracy but lower coverage than the Language use method~(see Section~\ref{sec:eval}).
\end{itemize}

\subsubsection{Country of residence}
The only published method for Reddit user localisation to date~\citep{GeocodingReddit_Harrigian_18} infers a user's country of residence via a dirichlet process mixture model\footnote{\url{https://github.com/kharrigian/smgeo}}. It uses the distribution of words, posts per subreddit, and posts per hour of the day~(timezone proxy) of a user's up to~250 most recent comments.

\subsubsection{Binary gender}
\label{sec:methods-gender}
Three methods to recognise binary gender~(feminine~(f)/masculine~(m)) leveraging different types of information were compared.
All three methods pertain to a performative gender view, which posits that people understand their and others' gender identity by certain behaviours~(including language) and appearances that society stipulates for bodies of a particular sex~\citep{GenderVariableNLP_Larson_17}.
Non-binary gender identities were not included due to a lack of NLP methods to detect them.

\begin{itemize}
\item \textbf{Username}: The character-based neural network model of~\citet{MeasuringSupportOnlineCommunities_Wang_18} predicts whether a username strongly performs f or m gender, otherwise it assigns no label.
\item \textbf{Self-reported}: See Section~\ref{sec:methods-age}.

\item \textbf{Language use}: The neural network model by~\citet{HAM_Tigunova_19} predicts gender for Reddit users with at least ten posts from the post texts. It was trained on data automatically labelled with self-reported gender provided by \citet{RedditUserTraits_Tigunova_20}~(see Appendix~\ref{sec:appendix-method-age+gender}).

\item \textbf{Hybrid}:
Evaluation 
revealed an accuracy ranking of Username > Self-reported > Language use and the inverse for coverage~(Section~\ref{sec:eval}).
The Hybrid method assigns a binary gender identity in a sequential approach, disregarding possible disagreements between methods: If the Username method found the username to perform f or m gender, it takes this prediction, otherwise assumes the self-reported gender if available, and else resorts to the predictions of the Language use method. 
\end{itemize}

\section{Ethical considerations}
At least four main ethical considerations arise for the work presented here: Concerns around (1) consent and (2) anonymity of Reddit users, 
around the (3) selection, category labels, and assignment of user characteristics~(MH diagnoses, age, country, gender), and (4) potentially harmful uses of the presented dataset and methods. 
The Lancaster University Faculty of Health and Medicine research ethics committee reviewed and approved this study in May~2019~(reference number FHMREC18066).

\subsection{Consent}
If and how research on social media data needs to obtain informed consent is debated~\citep{EthicalQualitativeInternetResearch_Eysenbach_01,ResearchSocialMediaUsersViews_Beninger_14,SocialMonitoringPublicHealth_Paul_17}, mainly because it is not straightforward to determine if posts pertain to a public or private context.
Legally, the Reddit privacy policy\footnote{\url{https://www.redditinc.com/policies/privacy-policy}} explicitly allows copying of user contents by third parties via the Reddit API, but it is unclear to what extends users are aware of this
~\citep{EthicsOnlineResearch_Ch4_TwitterDataSource_Ahmed_17}.
In practice it is often infeasible to seek retrospective consent from hundreds or thousands of social media users.
Current ethical guidelines for social media research~\citep{EthicalMentalHealthSocialMediaResearch_Benton_17,EthicalFrameworkPublishingTwitterData_Williams_17} and practice in comparable research projects~\citep{SuicidalityTwitter_ODea_15,EthicsOnlineResearch_Ch4_TwitterDataSource_Ahmed_17}, regard it as acceptable to waive explicit consent if users' anonymity is protected.
Therefore, Reddit users in this work were not asked for consent.

\subsection{Anonymity}
In line with guidelines for ethical social media health research~\citep{EthicalMentalHealthSocialMediaResearch_Benton_17}, this research only shares anonymised  and paraphrased excerpts from posts in publications.
Otherwise, it is often possible to recover usernames via a web search with the verbatim post text~(see also Section~\ref{subsec:datasharing}).

\subsection{Rationales for user characteristics}
As stated in the introduction, user characteristics are important to determine about which populations research on this dataset may generalise.
The NLP community increasingly expects data statements for datasets~\citep{DataStatementsNLP_Bender_18}, which include speaker age and gender specifications.
As Section~\ref{sec:results-demographics} shows, characteristics of Reddit users with a self-reported BD diagnosis deviate from both general Reddit user statistics and epidemiological studies, which therefore do not constitute useful proxies.
Relying entirely on self-reported information introduces selection biases because not all user groups may be equally inclined to explicitly share certain characteristics. 
This motivates using statistical methods to infer Reddit users' age, country, and gender here.

The user characteristics comorbid MH issues, age, country, and gender were chosen because they impact peoples' lived experience in BD as discussed in the following.
This work identifies users with a self-reported BD diagnosis because collecting posts from BD-specific subreddits does not suffice as carers and people who are unsure if they meet diagnostic criteria also post there.
Other self-reported MH diagnoses were extracted because people with BD diagnoses frequently experience additional MH issues~\citep{PrevalenceBP_Merikangas_11}.
Self-reported diagnoses capture only users who explicitly and publicly share their diagnosis. This research does not infer any users' MH state.

\citet{BDOlderAdults_Depp_04}, among others, provide evidence for age-related differences in BD symptoms and experiences, also through increasing importance of physical health comorbidities with ageing.
Age estimates were grouped in the same way as in a US survey of Reddit users for comparison.

Healthcare systems, including provision of MH care, vastly differ between countries, even within Western countries such as the US, UK, and Germany.
The MH services people can access may influence their experience of BD, motivating estimation of their country of residence.
While~\citet{GeocodingReddit_Harrigian_18} predicts longitude/latitude coordinates in 0.5~steps, these are mapped to countries because more fine-grained user localisations are not needed.

Using a gender variable in NLP deserves special consideration because it concerns people's identity~\citep{GenderVariableNLP_Larson_17}.
Biological sex can impact on the experience of BD, primarily through issues around childbirth and menopause, also related to mood-impacting hormonal changes~\citep{SexBD_Diflorio_10}; \citet{GenderIdentityBD_Sajatovic_11} found effects of gender identity on treatment adherence in BD.
This work only uses binary m/f gender labels since no NLP method with more diverse categories was available.
The gender recognition methods could cause harm to individual users if they were misgendered and then incorrectly addressed or referred to.
This project minimises such harm because the labels only serve to estimate the gender distribution and not to target individual users.

\subsection{Dual use}
This research aims to learn more about Reddit users who share their experiences with BD to yield findings that will ultimately lead to new or improved interventions that support living well with BD.
However, most research, even when conducted with the best intentions, suffers from the dual-use problem~\citep{ImperativeResponsibility_Jonas_84}, in that it can be misused or have consequences that affect people's life negatively. 
Adverse consequences of this study could arise for the Reddit users included in the dataset if they are sought out based on their self-reported BD diagnosis to be targeted with, e.g. medication advertisements.
The large number of Reddit posts in this dataset can serve as training data for machine learning systems that assign a likelihood to other Reddit/social media users for meeting BD criteria~\citep[e.g.,][]{SMHD_Cohan_18,BDPrediction_Sekulic_18}.
For example, health insurance companies could misuse this, using applicants' social media profiles in risk assessments.

\subsection{Transparency: Dataset and code release}
\label{subsec:datasharing}
Based on all above considerations, the dataset will only be shared with other researchers upon request and under a data usage agreement that specifies ethical usage of the dataset as detailed in this section.
The dataset release necessarily contains the original post texts but with replaced post and user ids.
This requires verbatim web searches with the post texts to seek out individual Reddit users and thus complicates automatisation and scaling.
User characteristics, including the manually annotated subsets, will only be shared separately with researchers who justify a specific need for them.
To aid transparency, the code and patterns to identify self-reported MH diagnoses, age, and gender are released\footnote{\url{https://github.com/glorisonne/reddit_bd_user_characteristics}}.

\begin{table}[tbh]
\centering
\begin{tabular}{L{1.7cm}R{0.9cm}R{1.7cm}l}
Variable & Users & Agreement (\%) & Labels (\%)\\
\hline
Self-rep. BD diag. & 100 & 97.0 & Yes: 97.0\\
&&&No: 3.0\\
\hline
Date of birth   & 116 & 99.1 & Date: 90.5\\
&&&?: 19.5\\
\hline
Country         & 100 & 90.0 & US: 46.0\\
&&&CA: 9.0\\
&&&GB: 8.0\\
&&&Other: 25.0\\
&&&?: 12.0\\
\hline
Gender          & 116 & 95.7 & F: 51.7\\
&&&M: 34.5\\
&&&Trans: 0.9\\
&&&?: 13.8\\
\end{tabular}
\caption{\label{tab:annotation} Number of users in manual annotation, raw annotator agreement, and label distributions after resolving disagreements in discussion (?: no label assigned due to lack of user-provided information on Reddit)
}
\end{table}
\begin{table*}[hbt]
\centering

\begin{tabular}{p{1.3cm}R{1.4cm}p{4.5cm}R{2.1cm}R{2cm}R{2cm}}
Variable & Users$^{test}$ & Method & Accuracy$^{test}$ & Coverage$^{test}$ & Coverage$^{all}$ \\
\hline
Age     & 105 &Self-reported  &	100.0\%	&98.1\% &11.5\%\\
group 	&       &Language use   & 60.6\%&	94.3\%  &66.0\%\\
		&       &Hybrid         &	99.0\%&	100\%   &68.3\%\\
\hline
Country &	88	& Words, subreddits, timing &78.4\%&	100\% & 100\%\\
\hline
Gender	&100	&Username       & 100\%	&12.0\% &10.9\%\\
		&       &Self-reported  &	97.9\%	&94.0\% & 11.9\%\\
		&       &Language use   & 84.2\%&	95.0\% & 66.0\%\\
		&       &Hybrid         &	97.0\%&	100\% &71.5\%
\end{tabular}
\caption{\label{tab:eval} Accuracy ($\frac{\text{correct}}{\text{total}}$) for user metadata extraction and inference methods (see Section~\ref{sec:methods-demographics}
) for manually annotated users (test), coverage ($\frac{\text{predicted}}{\text{total}}$) for manually annotated~(test) and all~(all, n=19,685) users
}
\end{table*}

\section{Results and discussion}
The self-reported BD diagnosis matching method identified 19,685 Reddit users who together had 21,407,595 public Reddit posts 
between~March~2006 
and~March~2019.
Compared to 9K~unique user accounts who posted in the four largest BD-related subreddits in May 2020, 
this likely only constitutes a small fraction of Reddit users with a BD diagnosis that could be reliably automatically identified (see following subsection).

\subsection{Manual annotation}
\label{sec:app-annotation}

Two authors manually annotated random subsets of users to evaluate all automatically extracted or inferred information according to the annotation guidelines\footnote{\url{https://github.com/glorisonne/reddit_bd_user_characteristics/blob/master/ManualAnnotationGuidelines.pdf}}.
As shown in Table~\ref{tab:annotation} agreement for all annotations was above 90\%, demonstrating feasibility and high reliability.

The annotators checked all extracted self-reported bipolar disorder diagnosis statements of 100 random included users, disagreeing only for three users (see first line of Table~\ref{tab:annotation})\footnote{No attempt was made to evaluate recall of user identification. Given an international prevalence of meeting BD criteria of about 2\%~\citep{PrevalenceBP_Merikangas_11} and expecting numbers of posts per account close to the average of 1,224 in the collected dataset, it was deemed infeasible to manually check all posts of randomly selected user accounts for self-reported bipolar disorder diagnosis statements.}.
The pattern matching approach for self-reported diagnosis statements mistakenly identified only three users (subsequently removed from the dataset) based on reports of other MH diagnoses where the word bipolar occurred close to the diagnosis term as well\footnote{Paraphrased excerpts of incorrectly identified self-reported BD diagnoses: \lq clinical depression with bipolar tendencies\rq, \lq diagnosed with BPD today, thought it was BD for years\rq, \lq diagnosed with depression, but sure I’ve got bipolar\rq.}.

To facilitate manual age and gender annotation, 116 users where randomly selected from the 2854~(14\%) of users where the Self-reported age or gender extraction method matched.
This explains the discrepancy between the coverage of the Self-reported method in Table~\ref{tab:eval} for the test set and full dataset.
The annotators only checked whether date of birth
or gender could be unambiguously extracted from all of a users’ posts that matched a self-reported age and gender pattern.
The test set for the gender evaluation results in Table~\ref{tab:eval} comprises only users labelled as m/f and excludes one manually identified transgender person.

\subsection{Evaluation of NLP methods}
\label{sec:eval}
Table~\ref{tab:eval} shows accuracy and coverage for the 
user characteristics extraction and inference methods described in Section~\ref{sec:methods-demographics} against the manually labelled users for which the annotators could determine a label.
%
For age, the Self-reported method outperforms the Language use method for accuracy but not coverage\footnote{The Language use method for age/gender does not have full coverage because it requires at least ten posts per user.
The methods agree for 62.6\% of the 1,788 users where both assign an age group.}.
The Hybrid method, subsequently used in Section~\ref{sec:results-age}, achieves 99\%~test set accuracy and 68\%~coverage on the full dataset. 
Harrigian's~(\citeyear{GeocodingReddit_Harrigian_18}) method assigns a country estimate to every user with 78\%~test set accuracy.
For gender, accuracy decreases from the Username, Self-reported, and Language use method, while coverage increases
\footnote{For 195 users where all three methods assign a gender identity, they agree on 73.8\% (90.8\% agreement between the Username and Self-reported method, 80\% between the Language use and Username or Self-reported method).}. 
The Hybrid gender identification method, used in Section~\ref{sec:results-age}, achieves 97\% test set accuracy, gender-labelling 72\%~of users..

\begin{table}[t!h]
\centering
\begin{tabular}{L{1.3cm}R{1.2cm}R{1.2cm}L{2cm}}
Diagnosis & Dataset\newline n=19,685 (\%) & SMHD\newline n=6,434 (\%)& Epidemiological studies (\%)\\
\hline
MDD	& 30.2 	&27.4&	N/A\\
\hline
Anxiety disorder	& 15.8 &	12.8&	13.3-16.8$^{*}$, n=921-1,537\\ 
\hline
ADHD	& 12.9&	9.6&	17.6$^{\dagger}$, n=399\\
\hline
BPD	& 8.4	&N/A	&16$^{\$}$, n=1,255\\
\hline
PTSD	& 6.5	&5.1	&10.8$^{*}$, n=1,185\\
\hline
OCD & 3.9	&3.4	&10.7$^{*}$, n=808\\
\hline
ASD	& 2.2	&2.0&	Unknown\\
\hline
ED	& 1.0	&0.8&	5.3-31$^{\odot}$, n=51-1,710\\
\hline
\end{tabular}
\caption{\label{tab:comorbidities} Self-reported comorbid diagnoses with BD in this work, the SMHD dataset, and epidemiological studies: $^{*}$\citet{BipolarAnxietyComorbidityEpidemiology_Nabavi_15}, $^{\dagger}$\citep{ADHDBDComorbidity_McIntyre_10},  $^{\$}$\citep{BPDBD_Zimmerman_13}, $^{\odot}$\citep{EatingDisoderBD_AlvarezRuiz_15}
}
\end{table}

\subsection{Reddit users’ characteristics}
\label{sec:results-demographics}
The following subsections compare characteristics of Reddit users with a self-reported BD diagnosis to general Reddit users and epidemiological statistics.

\subsubsection{Mental health comorbidities}
Table~\ref{tab:comorbidities} shows how many users disclosed other concurrent or lifetime MH diagnoses besides BD.
Rates for self-reported MH diagnoses in addition to BD are sightly higher in our dataset compared to the Self-reported MH diagnoses~(SMHD) dataset~\citep{SMHD_Cohan_18}, potentially because our dataset covers 27 more months of posts.

Like psychotic disorder (5.2\% of users prior to exclusion), a MDD diagnosis is mutually exclusive with BD according to the DSM~\citep[pp.~126,~134]{DSM-5_APA_13}\footnote{The dataset includes users with self-reported MDD but not psychotic disorder because depression but not psychosis is a core aspect of extreme mood, our focus of future research.}. 
A large part of identified self-reported MDD diagnoses were false positives where \lq depression\rq ~occurred near to a BD diagnosis statement. 
More conservatively only considering self-reported MDD diagnosis posts that do not also match BD patterns, results in 8.7\% users reporting both diagnoses.
MDD and Psychotic disorder diagnoses jointly with BD might indicate subsequently changed (mis-)diagnoses or disagreement of professionals.
Surveys in Germany~\citep{VersorgungserfahrungenBDDeutschland_Pfennig_11} and the US~\citep{MisdiagnosisBD_Hirschfeld_03} have shown that often more than ten years pass between onset of BD symptoms and receiving the diagnosis, with two thirds of people being misdiagnosed, most frequently with MDD.
Moreover, field trials for BD diagnoses with DSM-V criteria 
only showed moderate clinician agreement~\citep{FieldTrialsDSM-5_Freedman_13}. 

Comorbidity rates for anxiety disorders, BPD and PTSD align with results from epidemiological studies. 
Rates for comorbid ADHD, OCD, and ED are lower in the Reddit dataset population, which might in part be due to incomplete coverage of the patterns to capture diagnosis self-reports.
Additionally, epidemiological studies can be expected to yield higher comorbidity rates because they determine if participants meet criteria for various diagnoses with clinical interviews, whereas Reddit users may not have (or report) diagnoses for every condition they meet the criteria of.
Overall, 50.7\% of users reported at least one additional MH diagnosis, slightly less than three quarters of surveyed people in the World Mental Health Survey Initiative who met criteria for at least one other DSM-IV disorder besides BD~\citep{PrevalenceBP_Merikangas_11}.

More than 2\% of users reported an ASD diagnosis in addition to BD, with no epidemiological studies on ASD prevalence with BD yet. 
\citet{AutismBD_DellOsso_19} found significant levels of autistic traits among 43\%~of people with a BD diagnosis. 

\definecolor{b}{HTML}{4981CE}
\definecolor{g}{HTML}{859C27}
\definecolor{r}{HTML}{B22222}
\definecolor{o}{HTML}{FF6600}
\definecolor{darkgreen}{HTML}{228B22}

\pgfplotstableread{
j	dataset_first_post dataset_mean_posting_age reddit_US	
teenager	17.18	16.08   0.0
young	40.39	29.77    64
adult	37.60	47.54    29
middle	4.78	6.55    6
older	0.05	0.06    1
}\tableone

\begin{figure}[tb]
\centering
\begin{tikzpicture}
\begin{axis}[
    xbar,
    legend style={cells={anchor=west},at={(1.05,1.0),font=\small}, reverse legend},
    width= 1*\columnwidth, 
    axis y line*=left,
    axis x line*=bottom,
    label style={font=\footnotesize},
    tick label style={font=\footnotesize},
    minor x tick num=2,
    xtick = {0,10,20,30,40,50,60,70},
    xmin=0, 
    xlabel={Percentage},
    symbolic y coords={teenager, young, adult, middle, older},
    ytick={teenager, young, adult, middle, older},
    yticklabels={13-17,18-29,30-49,50-64,65+},
    area legend,
    bar width= 8pt,
    y=1.3cm, 
    enlarge y limits={0.15}, 
    nodes near coords={\pgfmathprintnumber[fixed,precision=1]\pgfplotspointmeta},
    nodes near coords align={horizontal}
]

\addplot [fill=b, draw=b, pattern color = b] table[x index=3,y index=0] {\tableone};
\addplot [fill=darkgreen, draw=darkgreen, pattern color = darkgreen, pattern = north west lines] table[x index=2,y index=0] {\tableone};
\addplot [fill=g, draw=g, pattern color = g, pattern = north east lines] table[x index=1,y index=0] {\tableone};

\legend{US Reddit users (n=288), {Dataset mean posting age}, {Dataset first post age (n=12,373)}}

\end{axis}  
\end{tikzpicture}
\caption{\label{fig:age}Age of Reddit users}
\end{figure}
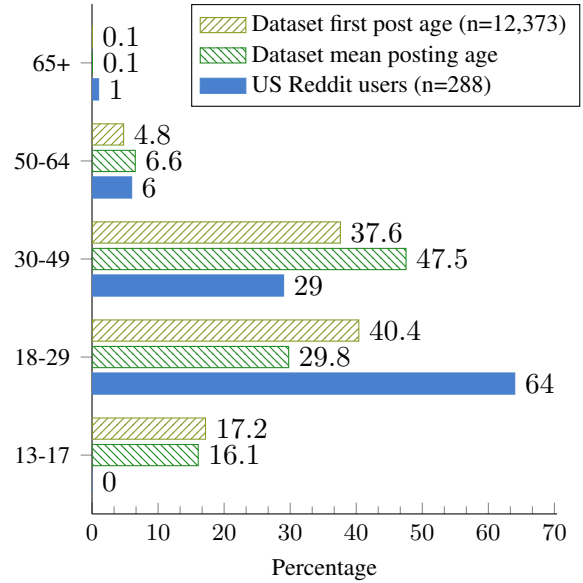
\subsubsection{Age}
\label{sec:results-age}
As shown in Figure~\ref{fig:age}, less Reddit users with a self-reported BD diagnosis are 18-29 but more 30-49 years old compared to average 
US Reddit users~\citep[p.~7]{RedditStatistics_PEW_16}\footnote{The \citet{RedditStatistics_PEW_16} survey only targeted adults, 
therefore there are no 13-17-year-old users.}.
The age of onset of BD symptoms is most frequently in late adolescence and early adulthood
~(\citeauthor{PrevalenceBDEurope_Pini_05}, \citeyear{PrevalenceBDEurope_Pini_05}; \citeauthor{PrevalenceBP_Merikangas_11}, \citeyear{PrevalenceBP_Merikangas_11}, p.~6).
In line with this, the majority of Reddit users who disclose a BD diagnosis are between 13-29 years old at their first post.
In the Global Burden of Disease study 2013, BD 12-months prevalence rates were significantly elated for 20-54 year olds~\citet[p.~447]{PrevalenceBurdenBD_Ferrari_16}.
In our dataset, almost 80\% of the Reddit users are 18-49 years old at their first post.

\subsubsection{Country of residence}
As shown in Table~\ref{tab:country}, more than 80\% of the Reddit users with a self-reported BD diagnosis are estimated to live in the US, and 95\% in one of the English-speaking countries US, UK, Canada, Australia.
This ranking aligns with site visitors of the Reddit desktop version~\citep{TrafficByCountryReddit_Statista_20}, although US users are even more prevalent in the BD dataset.
All of the top-5 countries in the dataset have a 12-months prevalence of BD diagnoses above the global average of 0.62\% according to the 2017 Global Burden of Disease Study~\citep{GBD_18}.

\begin{table}
\centering
\begin{tabular}{lrrr}
Country	&Dataset &Reddit.com &12-months\\
&(\%) &traffic (\%)	&prev. (\%)\\
\hline
US	    &81.9   &49.69  &	0.68\\
UK	    &5.6    &7.93   &	1.11\\
Canada	&4.9    &7.85   &	0.75\\
Australia &1.7	&4.32	&   1.15\\
Germany	&1.4 &	3.17	&   0.83\\
\end{tabular}
\caption{\label{tab:country} Top 5 estimated countries of residence of Reddit users with a self-reported BD diagnosis, location of reddit.com site visitors~\citep{TrafficByCountryReddit_Statista_20} and 12-months prevalence of BD~\citep{GBD_18}
}
\end{table}

\definecolor{b}{HTML}{4981CE}
\definecolor{g}{HTML}{859C27}
\definecolor{r}{HTML}{B22222}
\definecolor{o}{HTML}{FF6600}

\pgfplotstableread{
j	dataset	username	reddit_US
male	47.8	41.7	67.0
female	52.2	8.8	33.0 
}\tableone

\begin{figure}[tb]
\centering
\begin{tikzpicture}
\begin{axis}[
    xbar,
    legend style={cells={anchor=west},at={(1.0,1.65),font=\small}, reverse legend},
    width= 1*\columnwidth, 
    height=4cm,
    axis y line*=left,
    axis x line*=bottom,
    label style={font=\footnotesize},
    tick label style={font=\footnotesize},
    minor x tick num=2,
    xtick = {0,10,20,30,40,50,60,70},
    xlabel={Percentage},
    symbolic y coords={male, female},
    ytick={male, female},
    yticklabels={m, f},
    area legend,
    bar width= 8pt,
    enlarge y limits={0.5}, 
    nodes near coords={\pgfmathprintnumber[fixed,precision=1]\pgfplotspointmeta},
    nodes near coords align={horizontal}
]

\addplot [fill=b, draw=b, pattern color = b] table[x index=3,y index=0] {\tableone};
\addplot [fill=r, draw=r, pattern color = r, pattern = north west lines] table[x index=2,y index=0] {\tableone};
\addplot [fill=g, draw=g, pattern color = g, pattern = north east lines] table[x index=1,y index=0] {\tableone};

\legend{{US Reddit users (n=288)}, {General Reddit usernames (n=1,231,330)}, {Dataset (n=14,069)}}
\end{axis}  
\end{tikzpicture}
\caption{\label{fig:gender}Binary gender of Reddit users}
\end{figure}
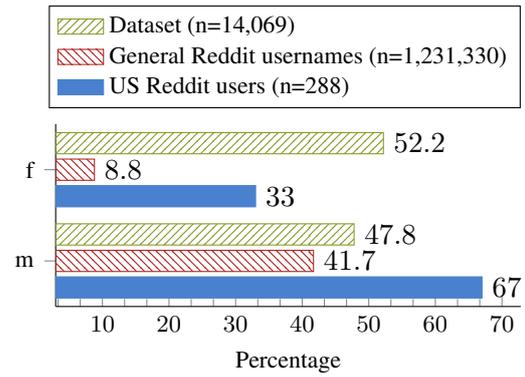

\subsubsection{Binary gender}
\label{sec:results-gender}
Figure~\ref{fig:gender} shows that the Hybrid method assigned feminine gender to slightly more than half of the Reddit users for which it ascribed a gender identity. 
This sharply contrasts with only 9\%~feminine vs. 41\%~masculine gender-performing usernames among Reddit users who posted in the top 10K subreddits with most posts~\citep{MeasuringSupportOnlineCommunities_Wang_18}.
A survey of adult US Reddit users~\cite{RedditStatistics_PEW_16} found that two thirds were men.

In epidemiological studies, biological men and women are equally likely to meet criteria for BD overall~(\citeauthor{PrevalenceBDEurope_Pini_05}, \citeyear{PrevalenceBDEurope_Pini_05}, \citeauthor{DSM-5_APA_13}, \citeyear{DSM-5_APA_13}, p.~124)\, although there is evidence that BD-II is more frequently diagnosed among women~\citep{SexBD_Diflorio_10}.
\citet{GenderIdentityBD_Sajatovic_11} found that biological men with a BD diagnosis scored significantly lower on masculine gender identity than the general male population, while there were no gender identity differences for biological women.
Considering a majority of male Reddit users and sex-equal prevalence of the diagnosis, feminine-gender-identifying people with a BD diagnosis seem to be more likely to use Reddit and/or to disclose their diagnosis.
The increased rates of female-gender identifying Reddit users with a self-reported BD diagnosis might also point towards a higher relative frequency of BD-II diagnoses (compared to BD-I) in this population.

\section{Limitations and implications}
\label{sec:discussion}

\subsection{Limitations}
First, unlike in clinical studies with face-to-face interactions, we cannot assume that every Reddit user in the dataset corresponds to one person. 
Additionally, self-reported diagnoses cannot be confirmed with diagnostic interviews as in clinical research.

Furthermore, there are several limitations to the NLP methods to infer user characteristics.
The method to extract self-reported MH diagnoses does not distinguish between actual comorbidities and misdiagnoses or previous diagnoses, for which symptoms may have resolved.
Manual evaluation of ten users with BPD comorbidity showed that seven reported concurrent diagnoses, one a BD to BPD change, one a BPD misdiagnosis, and one referred to BD by 
\lq BPD\rq.
Harrigian's~(\citeyear{GeocodingReddit_Harrigian_18}) method indicates the predominantly reflected country in a user's most recent posts, disregarding relocations.

The Self-reported age and gender extraction method is fallible to users providing incorrect information, for example disguising themselves as younger than they really are on dating subreddits.
Finally, none of the gender inference methods allow us to estimate how many users identify as transgender or non-binary.
Such indications were also too diverse to be captured in the regular expressions for self-reported age and gender.
Still, four of the subreddits with more than 10K posts by users with a self-reported BD diagnosis target transgender people, indicating that a proportion of the users in this research may not identify with their born sex.

\subsection{Health research implications}
Most importantly this work provides the first large-scale characterisation of Reddit users with a self-reported BD diagnosis, who are on average 27.7~years old at their first post, seem to overwhelmingly live in the US, and are more likely to identify with the feminine gender. Insofar they deviate from general Reddit as well as epidemiological statistics and also from participants in clinical studies. 

A large meta-analysis of psychological interventions for BD~\citep{PsychologicalInterventionsBDReview_Oud_18} showed that in
55 trials conducted across twelve countries~(35\% in the US) comprising 6,060 adults with BD, 89\% had recruited participants with a mean age higher than the 30~year-average of adult Reddit users with a self-reported BD diagnosis. 67\% of the trials recruited a higher percentage of females than the 52\% figure in the Reddit dataset~\citep[Table~DS2]{PsychologicalInterventionsBDReview_Oud_18}.
This cautions against generalising findings from Reddit data to all people with a BD diagnosis, but stresses its complementary role to clinical studies with different selection biases.

Another important implication is that NLP analysis of Reddit social media users largely confirmed high prevalence rates for comorbid MH conditions with BD from epidemiological studies.
Besides clinically established comorbidities with, e.g., Anxiety disorder and ADHD, the present analysis also revealed substantial prevalence of ASD, for which there is little clinical research to date.
Reddit may constitute a useful platform to learn about the experiences of people with BD with such currently under-researched comorbidities and may be a way to target them for recruitment to clinical studies.

\subsection{NLP research implications}
This work evaluated state-of-the-art methods to infer Reddit user characteristics~\citep{GeocodingReddit_Harrigian_18,MeasuringSupportOnlineCommunities_Wang_18,HAM_Tigunova_19} and demonstraed their utility in applied research.
A hybrid method achieved the best accuracy and coverage for age and gender identity by using high-accuracy information from self-reports ~(or a gender-performing username
) when available, filling in information for more users with less accurate predictions from a neural network language use-based method~\citep{HAM_Tigunova_19}.

Importantly, gender-inference methods so far are limited to detecting binary gender, although, e.g., 0.4\% of the US population identify as transgender~\citep{TransgenderPopulation_Meerwijk_19}.
Off-the-shelf NLP tools supporting a wider range of gender identities may be more inclusive and give more visibility to these groups of people in research. However, important ethical considerations arise around identifying people with transgender and non-binary gender identities, which are often stigmatised.

\section{Conclusion}
This paper set out to automatically profile Reddit users 
under consideration of ethical aspects.
A combination of pattern-based and previously published NLP methods served to estimate clinical, demographic, and identity characteristics of nearly 20K Reddit users with a self-reported BD diagnosis. 
Half of the Reddit users disclosed MH diagnoses besides BD and 80\% were located in the US.
From the users for which age or gender could be estimated, 80\% were between 18-49 years old and 52\% performed or identified with feminine gender.

These findings indicate about which populations BD-focused research on Reddit may generalise.
Additionally, this work may serve as a model for how to provide more information on other specific Reddit populations as requested by recent transparency and accountability movements in NLP.

\section*{Acknowledgements}
We would like to thank Anna Tigunova and Keith Harrigian for their assistance in applying their Reddit user profiling NLP tools.
We would also like to express our heartfelt thanks to Daisy Harvey, Stephen Mander, and the anonymous reviewers for helpful comments on a draft version of this article, to Andrew Moore for testing the code release, and to Alistair Baron for the initial idea for this work.

\bibliography{references}
\bibliographystyle{acl_natbib}

\appendix
\label{sec:appendix}

\section{Further method details}
\label{sec:appendix-method}
\subsection{Age and gender: Language use}
\label{sec:appendix-method-age+gender}
Tigunova et al.'s~(\citeyear{HAM_Tigunova_19}) HAM$_\text{CNN-attn}$ model predicts an age group\footnote{younger than 14, 14-23, 24-45, 46-65, 66+, relative to the user's most recent post} and gender for Reddit users with at least ten posts based on their up to 100 most recent posts.
Separate HAM$_\text{CNN-attn}$ models were trained on the RedDust dataset~\citep{RedditUserTraits_Tigunova_20} with the HAM open-source implementation\footnote{\url{https://github.com/Anna146/HiddenAttributeModels}} with the hyper-parameters specified by~\citet{RedditUserTraits_Tigunova_20}~(128 CNN filters of size 2, attention layer with 150 units, 70 training epochs).
Likely due to random seed variation, our trained age model had an area under the curve (AUROC) score of 0.80 compared to 0.88 in~\citet{RedditUserTraits_Tigunova_20}.
Our trained gender model had 84.9\% accuracy on the RedDust test set compared to 86.0\% reported by~\citet{RedditUserTraits_Tigunova_20}.

\subsection{Age: Hybrid method}
Two corrections were applied prior to the Hybrid method: The first author checked all users with a self-reported average posting age below 16 or above 60.
Age at account creation predictions younger than 13 by the Language use approach were discarded as Reddit requires an age of at least 13 when signing up.

\subsection{Country}
The Reddit country inference method~\citep{GeocodingReddit_Harrigian_18} initially was a proprietary project but later the first author, Keith Harrigian, rebuilt it for the public release\footnote{\url{https://github.com/kharrigian/smgeo}} used in this work.
Therefore, the training data and model performance, provided by Keith Harrigian in personal email communication on 5th March 2021, slightly differ from the original publication.
The training data consists of 56,853 automatically location-labelled users~(top 5: 68.8\%~US, 9.4\%~Canada, 7.0\%~UK, 3.3\%~Australia, 1.0\%~Germany), of which 8.2\% were identified based on self-reported locations in r/AmateurRoomPorn and the remainder by self-reported locations in reply to \lq Where are you from?\rq ~questions~\citep{GeocodingReddit_Harrigian_18}. 
Label precision was 97.6\% in a manual evaluation of 500 users\footnote{\url{https://github.com/kharrigian/smgeo\#dataset-noise}}.

The \lq Global\rq~(as opposed to US only) model was used to predict user locations, which achieves 35.6\% accuracy at 100 miles in 5-fold cross validation, equal to the originally reported performance in~\citet{GeocodingReddit_Harrigian_18}.
Overall country-level accuracy is 81.9\% and is generally higher for users with more training data~(95.1\% US, 65.1\% Canada, 82.8\% UK, 44.1\% Australia, 41.1\% Germany). 

\subsection{Gender: Username method}
\citet[p. 38]{MeasuringSupportOnlineCommunities_Wang_18} trained their long short-term memory~(LSTM) gender inference model on 80\% of 4,900,250 Twitter and 367,495 Reddit usernames, automatically labelled with self-reported m or f gender identity.
Following them, the present work assumes usernames to perform masculine~(m) gender for model predictions of~0.1 or lower, and feminine~(f) for 0.9 or higher.
This model and setting achieved 0.92 precision with 0.18 recall in 10\% held-out Twitter and Reddit username test data~\citep[Figure 5 in supplementary material]{MeasuringSupportOnlineCommunities_Wang_18}.

\begin{figure*}[t]
\includegraphics[scale=0.6]{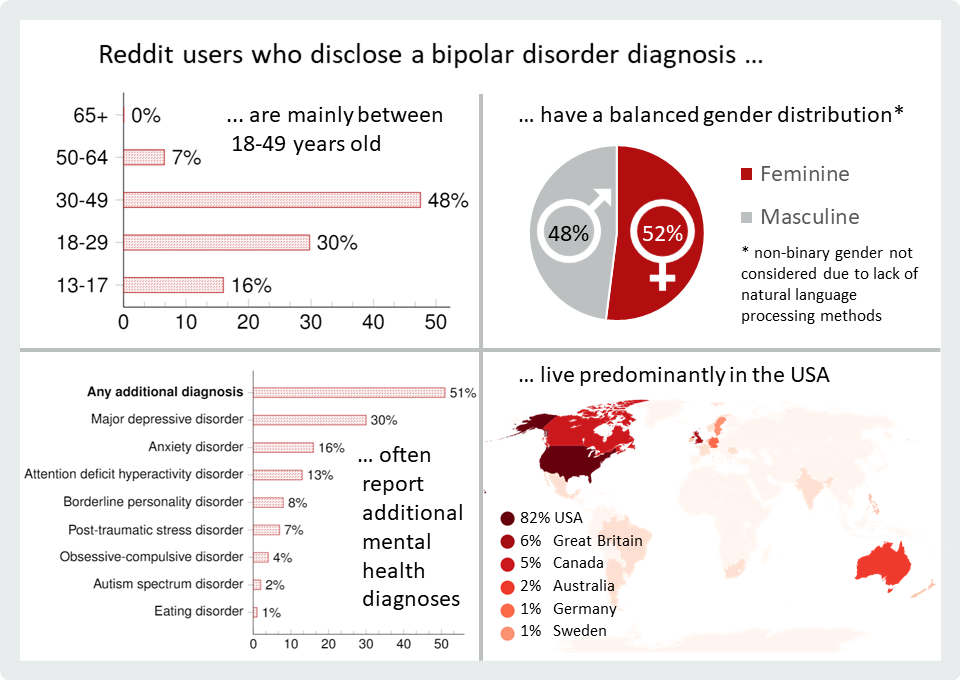}
\caption{\label{sec:appendix-visual-abstract} Visual summary of the characteristics of Reddit users who self-report a bipolar disorder diagnosis}
\end{figure*}

\end{document}